\documentclass[10pt,twocolumn,letterpaper]{article}

\usepackage{cvpr}
\usepackage{times}
\usepackage{epsfig}
\usepackage{graphicx}
\usepackage{amsmath}
\usepackage{amssymb}
\usepackage{bbm}
\usepackage{animate}
\usepackage{math_definitions}
\usepackage[ruled]{algorithm2e}
\usepackage[pagebackref=true,breaklinks=true,letterpaper=true,colorlinks,bookmarks=false]{hyperref}
\PassOptionsToPackage{bookmarks=false}{hyperref}

\cvprfinalcopy % *** Uncomment this line for the final submission

\def\cvprPaperID{****} % *** Enter the CVPR Paper ID here

\ifcvprfinal\fi
\def\trace{{\mathrm{trace}}}

\begin{document}

%%%%%%%%% TITLE
%\title{Cloud removal in temporal satellite image sequences \\ by
%smooth robust matrix completion}
\title{Removing Clouds and Recovering Ground Observations in Satellite
   \\ Image Sequences via Temporally Contiguous Robust Matrix Completion}

\if 0
\author[$\sharp$]{Jialei Wang}
\author[$\dag$]{Peder A. Olsen}
\author[$\dag$]{Andrew R. Conn}
\author[$\dag$]{Aur\'{e}lie C. Lozano}
\affil[$\sharp$]{University of Chicago jialei@uchicago.edu}
\affil[$\dag$]{IBM T.J. Watson Research Center}
\fi

\author{Jialei Wang\\
University of Chicago\\
{\tt\small jialei@uchicago.edu}
% For a paper whose authors are all at the same institution,
% omit the following lines up until the closing ``}''.
% Additional authors and addresses can be added with ``\and'',
% just like the second author.
% To save space, use either the email address or home page, not both
\and
Peder A. Olsen \quad\quad\quad Andrew R. Conn \quad\quad\quad Aur\'{e}lie C. Lozano\\
IBM T.J. Watson Research Center\\
{\tt\small \{pederao,arconn,aclozano\}@us.ibm.com}
}

\maketitle
%\thispagestyle{empty}

%%%%%%%%% ABSTRACT
\begin{abstract}
We consider the problem of removing and replacing clouds in satellite
image sequences, which has a wide range of applications in remote sensing.
 Our approach first detects and removes the
cloud-contaminated part of the image sequences. It then recovers the
missing scenes from the clean parts using the proposed ``\textsf{TECROMAC}'' (TEmporally Contiguous RObust MAtrix Completion) objective.  The objective function balances
temporal smoothness with a low rank solution while staying close to
the original observations.  The matrix whose the rows are pixels and
columns are days corresponding to the image, has low-rank because the pixels
reflect land-types such as vegetation, roads and lakes and there are
relatively few  variations as a result.  We provide efficient optimization algorithms
for \textsf{TECROMAC}, so we can exploit images containing millions of
pixels. Empirical results on real satellite image sequences, as well as
simulated data, demonstrate that our approach is able to recover underlying images
from heavily cloud-contaminated observations.
\end{abstract}

%%%%%%%%% BODY TEXT
\vspace{-0.5 cm}
\section{Introduction}
Optical satellite images are important tools for remote sensing, and
suitable analytics applied to satellite images can often benefit applications such
as land monitoring, mineral exploration, crop identification,
etc. However, the usefulness of satellite images is largely limited by
cloud contamination \cite{availability08ju}, thus a cloud removal
and reconstruction system is highly desirable.

In this paper, we propose a novel approach for cloud removal in
temporal satellite image sequences.  Our method improves upon exisiting cloud removal approaches \cite{cloud15huang,recover2014li,cloud13lin,missing13lorenzi,inpainting09maalouf,contextual06melgani,
  restoration09rakwatin} in the following ways:  1) the
approach does not require additional information such as a cloud mask,
or measurements from non-optical sensors; 2) our model is robust even
under heavy cloud contaminated situations, where several of the images could be
entirely covered by clouds; 3) efficient optimization algorithms makes
our approaches scalable to large high-resolution images.

We procede in two stages: a cloud detection stage and a
scene reconstruction stage. In the first, we detect clouds based
on pixel characteristics.\footnote {For example in 8-bit
  representations (0,0,0) is typically black while (255,255,255) is
  white, so clouds can be assumed to have high pixel intensity.} After
removing the cloud-contaminated pixels, we recover the background using a
novel ``\textsf{TECROMAC}'' (TEmporally Contiguous RObust MAtrix Completion) model, which
encourages low rank in time-space, temporal smoothness and robustness
with respect to errors in the cloud detector.  Our model overcomes the usual
limitations of traditional low-rank matrix recovery tools such as
matrix completion and robust principal component analysis, which
cannot handle images covered entirely by clouds.  We provide efficient
algorithms, which are based upon an augmented Lagrangian method (ALM)
with inexact proximal gradients (IPG), or alternating minimization (ALT), to handle the challenging non-smooth optimization problems related to \textsf{TECROMAC}. Empirical
results on both simulated and real data sets demonstrate the
efficiency and efficacy of the proposed algorithms.

%% The paper is organized as follows: Section \ref{sec:related_work}
%% discusses existing related work. We present our cloud removal
%% procedure in Section \ref{sec:procedure}, and derive the efficient
%% optimization algorithm in Section \ref{sec:optimization}. We
%% demonstrate the performance of our approach via simulation in Section
%% \ref{sec:simulation} and on real data sets in Section
%% \ref{sec:real_data}.  Section \ref{sec:conclusion} concludes with a
%% brief summary and discussion of future research directions.

\begin{figure*}[t]
\begin{center}
\includegraphics[width = \textwidth]{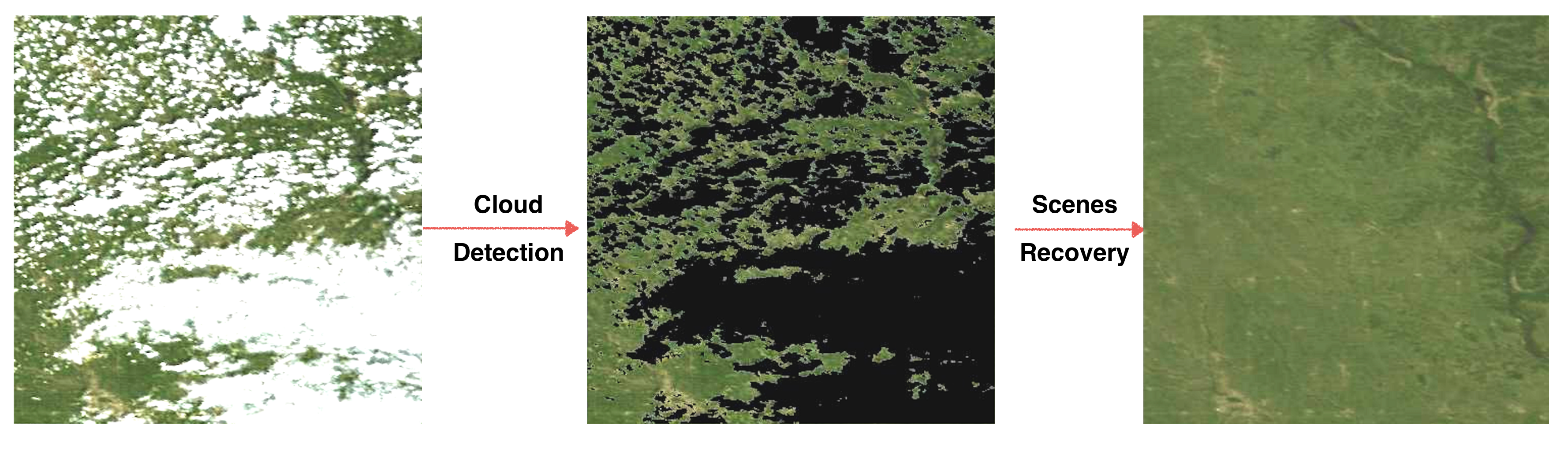}
\end{center}
\caption{Illustration of the proposed procedure: first stage: cloud detection and removal, second stage: background recovery.}
\label{fig:procedure}
\end{figure*}
%-------------------------------------------------------------------------

\section{Related work}
\label{sec:related_work}
Cloud detection, removal and replacement is an essential prerequisite
for downstream applications of remote sensing.  This paper belongs to
the important line of research of \emph{sparse representation}, which
has received considerable attention in a variety of areas, including
noise removal, inverse problems, and pattern
recognition~\cite{Elad10,Rubinstein10,Wright10}.  The key idea of
sparse representation is to approximate signals as a sparse
decomposition in a dictionary that is learnt from the data
(see~\cite{Mairal14} for an extensive survey).  For the task of cloud
removal and image reconstruction, \cite{missing13lorenzi} developed
compressive sensing approaches to find sparse signal representations
using images from two different time points.  Subsequently \cite{recover2014li}
extended dictionary learning to perform multitemporal recovery from
longer time series.  A group-sparse representation approach was
recently proposed by~\cite{Li15} to leverage both temporal sparsity
and non-local similarity in the multi-temporal images. 
A method was proposed in \cite{recover2014li} to co-train two dictionary pairs, one pair generated from
the high resolution image (HRI) and low resolution image
(LRI) gradient patches, and the other generated from the
HRI and synthetic-aperture radar (SAR) gradient patches.
It is demonstrated that such a combination of multiple data
types improves reconstruction results as it is able to provide
both low- and high-frequency information.

Our method has its origin in robust principal component analysis,
\cite{robust11candes,de2001robust} and matrix completion \cite{candes2009exact,power10candes}. However, these models require uniform or weighted sampled observations to ensure the recovery of low-rank matrix  \cite{foygel2011learning,negahban2012restricted,DBLP:conf/nips/SalakhutdinovS10}, and thus cannot handle images with extensive cloud
cover.

%-------------------------------------------------------------------------
\section{Problem description and approach}
\label{sec:procedure}
Let $\Ycal \in \RR^{m \times n \times c \times t}$ be a 4-th order
tensor whose entry values include the intensity and represent the observed cloudy satellite image sequences. The dimensions corresponds to lattitude, longitude,
spectral wavelength (color) and time.  The images have size $m\times
n$ with $c$ colors\footnote {A standard digital image will have a red,
  green and blue channel. A grayscale image has just one channel} and
there are $t$ time points corresponding to each image.  Our goal is to
remove the clouds, and recover the background scene, to facilitate
applications such as land monitoring, mineral exploration and
agricultural analytics.

As already mentioned, our cloud removal procedure is divided into two stages: namely
\textbf{cloud detection} followed by the \textbf{underlying scene
  recovery}. The cloud detection stage aims to provide an indication
as to what pixels are contaminated by clouds at a particular time.
Given the cloud mask, the recovery stage attempts to recover the
obfuscated scene by leveraging the partially observed, non-cloudy
image sequences. Figure~\ref{fig:procedure} provide an illustration of
the proposed two-stage framework. We describe the details of these two
stages in the following sections.

\subsection{Detecting clouds}
\begin{figure}[t]
\begin{center}
\includegraphics[width = 0.5 \textwidth]{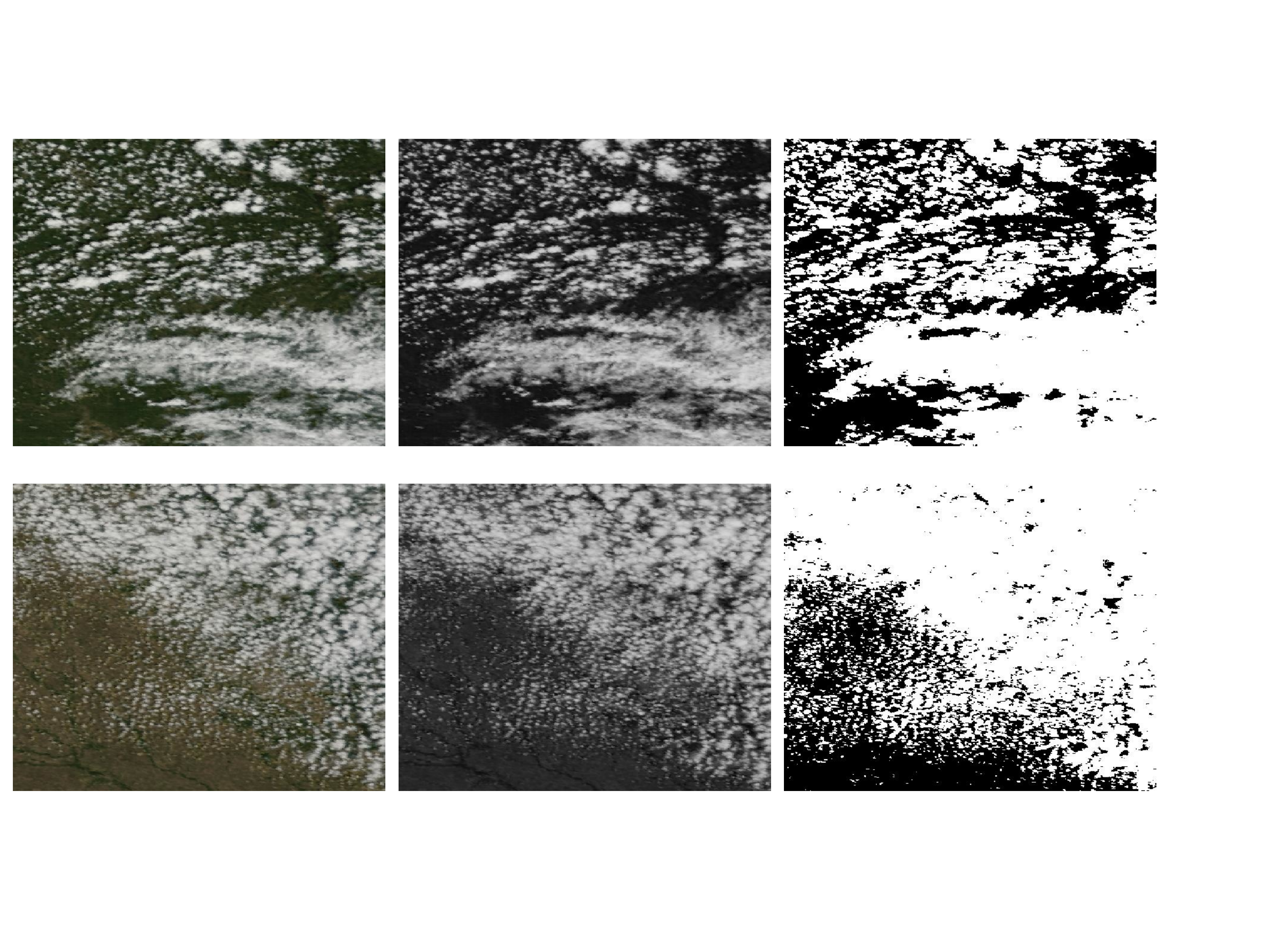}
\end{center}
\vspace{-0.5 cm}
\caption{Illustration of the proposed cloud detection procedure for
  two instances. Left: original image, middle: dark channel image, right: thresholded binary image.}
\label{fig:cloud_detection}
\end{figure}
Given the observed image sequences $\Ycal$, the goal of a cloud
detector is to produce a set of indices $\Omega$, called a cloud mask,
such that a lattitude-longitude-color-time position
$(i,j,k,l)\in\mathbb{Z}_m\times\mathbb{Z}_n\times\mathbb{Z}_c\times\mathbb{Z}_t$
is covered by a cloud if $(i,j,k,l)\not\in\Omega$ and is not cloud
contaminated if $(i,j,k,l)\in\Omega$.  Note that typically the
entire color channel will be contaminated by clouds.

We design a simple, yet effective cloud detector by thresholding on
the dark channel (the minimum pixel intensity values across the RGB
channels), \cite{He:11} followed by a post-processing stage to
distinguish between a stationary “white background” and “white
clouds”.  This works because clouds are predominantly white so
that the minimum intensity value is large and there are typically no
other white objects on the ground (except for snow in winter, please
refer to section \ref{section:post_processing} below for a related
discussion).  Our estimator for $\Omega$ is therefore
\begin{align}
\hat \Omega = \{(i,j,k,l) | \min_{c = \{r,g,b\}} \Ycal(i,j,c,l) < \gamma \},
\label{eqn:thresholding}
\end{align}
where $\gamma$ is the thresholding value that controls the trade-offs between false positives/negatives. Figure \ref{fig:cloud_detection} illustrates the above described method.% with thresholds of ?,? and ?.

There are several other approaches for cloud detection, for example, one could  use support vector machines \cite{support95corinna} to build linear classifiers based on the RGB values. However, this approach requires additional labeled training data, which is usually not easy to obtain.  Moreover, in our experience we observed that the classification approach is typically no better than the simple thresholding method.% [ARC  How extensive is our experience? Enough?]

\subsubsection{Post processing after thresholding}
\label{section:post_processing}
The thresholding approach described above cannot distinguish between a
stationary ``white background'' and ``white clouds''.  White objects
in the landscape, such as houses, should remain in the reconstructed
images.  Fortunately, we can exploit the transient nature of clouds to
label pixels always absent from $\hat \Omega$ as having a white
background.  Since the stationary pixels are also contaminated by
clouds we maintain the background by treating the median pixel value
as the background color and locate some cloud free instances by a
k-nearest neighbors search \cite{Altman:92}.  This approach is primitive,
but since the stationary white objects do not dominate the scene, the
method is sufficient (though it will not be able to handle extreme cases such as heavy snow as background). 

The overall procedure for cloud detection is summarized in Algorithm
\ref{alg:cloud_detection} below.

\begin{algorithm}[hpt]
\textbf{Input}: temporal sequence of cloudy images $\Ycal$;\\
\quad parameter $K$ in k-nearest-neighbors search. \\
\textbf{Output}: $\Omega$: indication set of non-cloudy pixels.\\
\SetAlgoLined
1, Obtain the initial guess via thresholding as in \eqref{eqn:thresholding}, \\
\qquad $\Omega = \hat \Omega$\;
2, Identify the ``always white'' pixel sequences: \\
\qquad $\Wcal = \{(i,j)| \forall k,l\quad (i,j,k,l) \not\in \hat \Omega \}$\;
3, \For{$(i,j) \in \Wcal$}{
Compute the median pixel vector: \\
\qquad $\mb = \textsf{Median}(\{\Ycal(i,j,:,l),l \in [t]\})$\;
Find the indices of the k-nearest-neighbors: \\
\qquad $\tilde \Omega = \textsf{Knn-Search}(\{\Ycal(i,j,:,l),l \in [t]\},\mb,K)$\;
$\Omega \leftarrow \Omega \bigcup \tilde \Omega$.
}
\caption{Cloud detection procedure.}
\label{alg:cloud_detection}
\end{algorithm}

\subsection{Image sequences recovery}
Given the cloud detection result $\Omega$, the image sequences
recovery model reconstructs the background from the partially observed
noisy images. For pixels in $\Omega$ the recovered values should stay close
to the observations.  Also, the reconstruction should take the following key
assumptions/intuitions into consideration:
\begin{itemize}
\item \textbf{Low-rank}: The ground observed will consist of a few
  slowly changing land-types such as forrest, agricultural land, lakes
  and a few stationary objects (roads, buildings, etc.).  We can
  interpret the land-types as basis elements of the presumed low-rank
  reconstruction.  More complex scenes would thus require a higher
  rank, but we expect in general the number of truly independent
  frames (pixels evolving over time) to be relatively small.
\item \textbf{Robustness}: Since the cloud detection results are not
  perfect, and there is typically a large deviation between the cloudy
  pixels and the background scene pixels, the recovery model should be
  robust with respect to the (relatively) sparsely corrupted observations.
\item \textbf{Temporal continuity}: The ground contains time-varying objects
  with occasional dramatic changes (e.g. harvest).  Most of the time
  the deviation between two consecutive frames should not be large
  given appropriate time-steps (days or weeks).
\end{itemize}
{\bf Notation:} 
Let the observation tensor be reshaped into a matrix $Y \in \RR^{mn
  \times ct}$ defined by $Y_{uv}=\mathcal{Y}_{ijkl}$ when $u=i+jm$ and
$v=k+lc$.  Similarily, let $\mathcal{X}$ be the reconstruction we are
seeking and $X \in \RR^{mn
  \times ct}$ be the corresponding reshaped matrix.  Further we let
$\mathbb{I}$ denote the indicator function. Based on above discussed principles we propose the following
reconstruction formulation:
\begin{align*}
\min_{X}& \quad {\rm rank}(X) \\
\text{s.t.}& \quad \sum_{(i,j,k,l) \in \Omega} \mathbb{I}\biggl(\Xcal(i,j,k,l) \neq \Ycal(i,j,k,l)\biggl) \leq \kappa_1, \\
& \sum_{i=1}^m \sum_{j=1}^n\sum_{k=1}^c\sum_{l=2}^t \biggl(\Xcal(i,j,k,l) - \Xcal(i,j,k,l-1)\biggr)^2 \leq \kappa_2. \\
 \end{align*}
The above formulation finds a low-rank reconstruction $X$ such that:
\begin{itemize}
\item $X$ disagrees with $Y$ at most $\kappa_1$ times in the predicted
 non-cloudy set $\Omega$, but can disagree an arbitrary amount on the
  $\kappa_1$ pixels (robustness);
\item The sum of squared $\ell_2$ distances between two consecutive frames is bounded by $\kappa_2$ (continuity).
 \end{itemize}

Unfortunately, both the rank and the summation of the indicator function values are non-convex,
making the above optimization computationally intractable in practice, thus we  introduce our temporally contiguous robust matrix completion (\textsf{TECROMAC}) objective which is computationally efficient, first introducing the forward
temporal shift matrix, $S_+ \in \RR^{t \times t}$ defined as
\begin{align*}
[S_+]_{i,j} = \begin{cases}
1 &\text{if } i = j + 1, i < t \\
1 &\text{if } i = j = t \\
0 &\text{otherwise}
\end{cases}.
\end{align*}
and the discrete derivative matrix $R \in \RR^{ct \times t}$ as
\begin{align*}
R = \begin{bmatrix}
I_t - S_+ \\
I_t - S_+ \\
\ldots \\
I_t - S_+
\end{bmatrix}
\end{align*}
Let $\|\cdot\|_F$ denote the Frobenius norm, the \textsf{TECROMAC} objective can be written as:
\begin{align}
\min_{X}  \norm{P_{\Omega}(Y-X)}_1 + \lambda_1 \norm{X}_* + \frac{\lambda_2}{2} \norm{XR}_F^2,
\label{eqn:TECROMAC}
\end{align}
where $\lambda_1$ controls the rank of the solution and $\lambda_2$
penalizes large temporal changes.

In the objective \eqref{eqn:TECROMAC}, the first term $\norm{P_{\Omega}(Y-X)}_1$ controls the reconstruction error on
the predicted non-cloudy set $\Omega$.  The second term encourages low
rank solutions, as the nuclear norm is a convex surrogate for direct
rank penalization \cite{DBLP:conf/colt/SrebroS05}.  Noting that
\begin{align*}
\norm{XR}_F^2= \sum_{i=1}^m\sum_{j=1}^n\sum_{k=1}^c\sum_{l=2}^t (\Xcal(i,j,l,k) - \Xcal(i,j,l,k-1))^2,
\end{align*}
is the finite difference approximation of the temporal derivatives one
can see that the last term encourages smoothness between consecutive
frames.

\subsection{Existing reconstruction model paradigms}
\begin{figure}
\begin{center}
\includegraphics[width = 0.5 \textwidth]{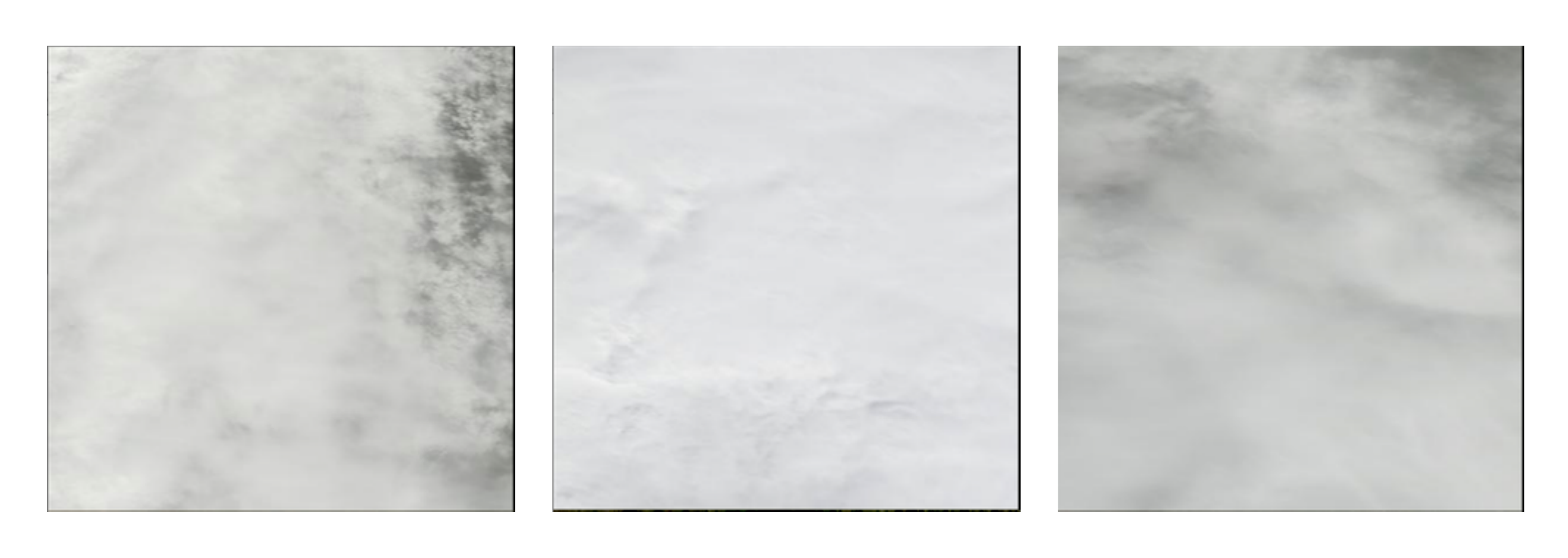}
\end{center}
\caption{Examples of totally cloud-contaminated frames.}
\label{fig:cloud_example}
\end{figure}

Following the cloud detection stage, there are several existing models
reconstruction paradigms that could be applied.
For example, one could use 
\begin{itemize}
\item Interpolation \cite{chronology02meijering}; although this violates the low-rank and robust assumptions
\item Matrix completion (MC) \cite{power10candes}; which violates
  robustness and continuity.
\item Robust matrix completion (RMC) \cite{lowrank13chen}; which uses $\norm{\cdot}_1$ loss instead of $\norm{\cdot}_F^2$ to ensure robustness. However, it still violates the
  continuity assumption.  RMC is
  an extension to MC and is inspired by robust principal component
  analysis (RPCA) \cite{robust11candes},  and has been analyzed theoretically \cite{lowrank13chen,DBLP:conf/icml/ChenXCS11}.
\end{itemize}
MC and RMC both require each image to have some pixels not corrupted
by clouds to ensure successful low-rank matrix recovery \cite{foygel2011learning,negahban2012restricted,DBLP:conf/nips/SalakhutdinovS10}.  Unfortunately, this is often not the case, as seen in
Figure~\ref{fig:cloud_example}. Tropical regions for example can be
completely covered by clouds as often as 90\% of the time.
Consequently, as we also show in the experiments below, directly
applying MC or RMC will lead to significant information loss
for the heavily contaminated frames.  Table~\ref{table:comparison}
summarizes the properties of the different approaches.
\begin{table}[ptbh]
\begin{center}
\begin{tabular}{c | c | c | c | c}
			Approach& Interpolation &MC & RMC & \textsf{TECROMAC} \\
			\hline \hline
			Low-rank & $\times$ & \checkmark & \checkmark  & \checkmark  \\
			Robustness    & $\times$ & $\times$  & \checkmark  & \checkmark  \\
			Continuity   & \checkmark & $\times$  & $\times$  & \checkmark  \\
			\hline
\end{tabular}
\end{center}
\caption{Comparison of properties of various approaches.}
\label{table:comparison}
\end{table}

One could also use RPCA \cite{robust11candes} or stable principle component pursuit (SPCP) \cite{DBLP:conf/isit/ZhouLWCM10} directly on cloudy image sequences \cite{DBLP:conf/uai/AravkinBCO14}, without the cloud detection stage. However, these approaches only work for images with sparse cloud, and usually fail under more challenging dense cloudy case, as we have observed in practice. 

%-------------------------------------------------------------------------
\section{Optimization algorithms}
\label{sec:optimization}

In this section, we present computationally efficient algorithms to
solve the optimization problem \eqref{eqn:TECROMAC}.  The optimization is
challenging since the objective \eqref{eqn:TECROMAC} contains two
non-smooth terms (the $\ell_1$ term and the nuclear norm term).  Our
overall algorithms are based on an augmented Lagrangian methods (ALM)
\cite{constrained82bertsekas,lin2010augmented,nocedal2006numerical}, with special designs for the subproblem
solvers.

First, we re-write \eqref{eqn:TECROMAC}
in the equality constrained form:
\begin{align}
\min_{E,X} &  \norm{P_{\Omega}(E)}_1 + \lambda_1 \norm{X}_* + \frac{\lambda_2}{2} \norm{XR}_F^2. \\
&\text{s.t. } Y = X+E.
\label{eqn:TECROMAC_eq}
\end{align}
The augmented Lagrangian method (ALM) tackles the
inequality constraints indirectly by introducing Lagrange
multipliers and a quadratic penalty to control deviations from
the equality constraint.  The unconstrained objective in our case is
\begin{align*}
L(X,E,Z,\mu) = &\norm{P_{\Omega}(E)}_1 + \lambda_1 \norm{X}_* + \frac{\lambda_2}{2} \norm{XR}_F^2  \\
&+ \langle Z, Y - X - E \rangle + \frac{\mu}{2} \norm{Y - X - E}_F^2,
\end{align*}
where $Z$ are the Lagrange multipliers corresponding to the equalities and $\mu$ is the quadratic penalty
weight.  A quadratic penalty function alone, requires that the penalty
parameter $\mu$ tends to infinity whereas the Lagrangian term allows
one to update the Lagrangian multipliers, thereby enabling the penalty
parameter to remain finite. 
The general framework of ALM works as follows: at each
iteration $r$ we solve the following problem:
\begin{align}
\{X^{r+1}, E^{r+1}\} = \arg\min_{X,E} L(X,E,Z^{r},\mu^r).
\label{eqn:sub_problem}
\end{align}
Then we update the dual variables $Z$ as
\begin{align}
Z^{r+1} = Z^r + \mu^r(Y-X^{r+1}-E^{r+1}),
\label{eqn:update_dual}
\end{align}
and also\footnote {Mathematical optimizers would only update one or the other, depending on conditions of feasibility versus stationarity, see for example \cite{Conn:91}} increase the penalty parameter $\mu$ via
\begin{align}
\mu^{r+1} = \rho \mu^r.
\label{eqn:update_mu}
\end{align}
The update for the multipliers, \eqref{eqn:update_dual}, can be
determined by differentiating the augmented Lagrangian with respect
to $Z$ and comparing coefficients with a Lagrangian form.

The augmented Lagrangian method is often efficient and robust, however, each iteration involves solving the subproblem \eqref{eqn:sub_problem}.  We describe two inexact algorithms to accelerate the optimization process:  the inexact proximal gradient (IPG), and the alternating minimization (ALT) algorithms, both of which use shrinkage operators \cite{bach2012optimization,DBLP:journals/ftopt/ParikhB14}.

%% which is
%% usually prohibitive for extremely large problems. One common approach for solving \eqref{eqn:sub_problem} in the latter context is to use an alternating minimization (or block coordinate descent) strategy to update
%%  $X$ and $E$ separately. However, in the most obvious approach this involves many singular value decompositions (SVD) of  very large matrices, which is computationally expensive. In the subsection,
%%  we describe two inexact algorithms to speed up the optimization process:  the inexact proximal gradient (IPG), and the alternating minimization (ALT) algorithms.

\subsection{Shrinkage Operators}
We write $S_\lambda(X)$ for the elementwise soft-thresholding
operators, $[S_\lambda(X)]_{ij}=\mathrm{sign}(x_{ij})(|x_{ij}|-\lambda)_+$ and
the singular value thresholding $\textsf{SVT}_\eta(X)=US_{\eta}(\Sigma)V^\top$, where $X=U\Sigma V^\top$ is the singular value decomposition
of $X$.

\subsection{Inexact Proximal Gradient}
\begin{algorithm}[hptb]
\textbf{Input}: Observation set $\Omega$ and data $Y$;\\
\quad regularization parameters $\lambda_1,\lambda_2$. \\
\textbf{Output}: Recovered matrix $X$.\\
\SetAlgoLined
\While{not converged}{
Updating $X$ using \eqref{eqn:ipg_x} \;
Updating $E$ using \eqref{eqn:ipg_e} \;
Updating $Z$ using  \eqref{eqn:update_dual} \;
Updating $\mu$ using \eqref{eqn:update_mu}.
}
\caption{ALM-IPG: Inexact Proximal gradient method for optimizing \eqref{eqn:TECROMAC_eq}.}
\label{alg:opt_ipg}
\end{algorithm}

In the inexact proximal gradient algorithm (which is summarized in
Algorithm \ref{alg:opt_ipg}), we only update $X$ and $E$ once to
avoid many expensive SVD computations. Given the current
$E^r,Z^r,\mu^r$, we minimizing the following objective with respect
to $X^r$:
\begin{small}
 \begin{align*}
   L(X) =&  \ \lambda_1\; \norm{X}_* + f(X), \text{where}\\
   f(X) =&  \frac{\lambda_2}{2} \norm{XR}_F^2 + \langle Z^r, Y - X - E^r \rangle \\
   &+ \frac{\mu^r}{2} \norm{Y - X - E^r}_F^2\\
 \end{align*}
\end{small}
Since $f(X)$ is quadratic we can also write it in the following forms
\begin{eqnarray*}
f(X) &=& \frac{1}{2}\trace(XHX^\top)+\trace(G^\top X)+f(0)\\
     &=& \frac{1}{2}\|XA-B\|_F^2+f(0)-\|B\|_F^2,
 \end{eqnarray*}
where the Hessian at 0 is represented by $H=\lambda_2RR^\top+\mu I$ and the gradient at 0 is $G=-Z-\mu(Y-E)$, $A A^\top=H$ and $B=-GA^{-\top}$.

Introducing the auxilliary function
$$
Q(X,\hat{X}) = F(X)+\frac{c}{2}\|X-\hat{X}\|_F^2-\frac{1}{2}\|(X-\hat{X})A\|_F^2
$$
we see that $Q(X,X)=F(X)$ and $Q(X,\hat{X})\geq F(X)$ if $cI\succ AA^\top=H$.
We can rewrite $Q(X,\hat{X})$ to clarify how we can efficiently optimize the function with respect to $X$.  We have
\begin{align*}
  Q(X,\hat{X})= &\lambda_1\|X\|_*+\frac{c}{2}\|X\|_F^2 \\
  & -\trace\left( X(AB^\top-AA^\top\hat{X}+c\hat{X})\right) + \mathrm{const}\\
\end{align*}
Completing the square gives
$$
Q(X,\hat{X})= \lambda_1\|X\|_*+\frac{c}{2}\|X-V\|_F^2  + \mathrm{const}\\
$$
where
$$
V=\hat{X}+\frac{1}{c}(AB^\top-AA^\top\hat{X})=\hat{X}-\frac{1}{c}\nabla f(\hat{X}).
$$
It follows that the optimum is achieved for
\begin{align}
X= \textsf{SVT}_{\lambda_1/c}\left(\hat{X}-\frac{1}{c}\nabla f(\hat{X})\right).
\label{eqn:ipg_x}
\end{align}
Since the optimum with respect to $\hat{X}$ is given by $\hat{X}=X$ this gives the iteration scheme
$$
X^{r+1}=\textsf{SVT}_{\lambda_1/c}\left(X^r-\frac{1}{c}\nabla f(X^r)\right).
$$
\begin{align*}
\nabla f(X^r) = \lambda_2 X^r RR^\top - Z^r - \mu^r(Y - X^r - E^r).
\end{align*}

%%  \begin{align*}
%%  \min_{X} &\ \lambda_1\; \norm{X}_* + \frac{\lambda_2}{2} \norm{XR}_F^2 + \langle Z^r, Y - X - E^r \rangle \\
%%  &+ \frac{\mu^r}{2} \norm{Y - X - E^r}_F^2
%%  \end{align*}
%% Recall that for a minimization problem on the form $\min_X \|X\|_* +
%% \frac{\eta}{2}\|X-Y\|_F^2$ the solution is the singular
%% value thresholding $SVT_{\eta}(Y)$ which can be expressed in terms of
%% the SVD of $Y=U\Sigma V^\top$ as 
%%  \begin{align*}
%%  \textsf{SVT}_{\eta}(X) = U(\Sigma - \eta I)_+ V^\top\!\!\!, 
%%  \text{ where }x_+ = \max(x,0).
%%  \end{align*}
%% Define the quadratic term of the IPG objective as 
%% $f(X) = \frac{\lambda_2}{2} \norm{XR}_F^2 + \langle Z^r, Y - X - E^r
%% \rangle + \frac{\mu^r}{2} \norm{Y - X - E^r}_F^2$.  Following
%% \cite{Cai:10} we take
%% a proximal gradient step to obtain $X^{r+1}$ as:
%%  \begin{align}
%%  X^{r+1} = \textsf{SVT}_{\eta \lambda_1} (X^r - \eta \nabla f(X^r)).
%%  \label{eqn:ipg_x}
%%  \end{align}
%%  Where $\eta$ is the learning rate, and the gradient of $f(\cdot)$ is
%%  \begin{align*}
%%  \nabla f(X^r) = \lambda_2 X^r RR^\top - Z^r + \mu^r(Y - X^r - E^r).
%%  \end{align*}
%
The objective with respect to $E$ is:
\begin{align*}
\min_E &\norm{P_{\Omega}(E)}_1 +  \langle Z^r, Y - X^{r+1} - E \rangle \\
&+ \frac{\mu^r}{2} \norm{Y - X^{r+1} - E}_F^2,
\end{align*}
with the closed form solution
\begin{align}
E_{ij} = \begin{cases}
\Scal_{1/\mu} (Y_{ij} - X^{r+1}_{ij}  + \frac{1}{\mu^r} Z^r_{ij}) &\text{if } (i,j) \in \Omega \\
Y_{ij} - X^{r+1}_{ij}  + \frac{1}{\mu^r} Z^r_{ij} &\text{if } (i,j) \not\in \Omega.
\end{cases}
\label{eqn:ipg_e}
\end{align}
%where $\Scal_{\lambda}(x) = \sgn(x) (|x| - \lambda)_+$.

 \subsection{Alternating Minimization}
 \begin{algorithm}[t]
\textbf{Input}: Observation set $\Omega$ and data $Y$;\\
\quad regularization parameter $\lambda_1,\lambda_2$, stepsize $\eta$. \\
\textbf{Output}: Recovered matrix $X$.\\
\SetAlgoLined
\While{not converged}{
\While{not converged}{
Updating $U$ using \eqref{eqn:alt_u} \;
Updating $V$ using \eqref{eqn:alt_v} \;
Updating $E$ using \eqref{eqn:ipg_e} \;
}
Updating $Z$ as  \eqref{eqn:update_dual} \;
Updating $\mu$ as \eqref{eqn:update_mu}.}
\caption{ALM-ALT: Alternating minimization method for optimizing \eqref{eqn:TECROMAC_eq}.}
\label{alg:opt_alt}
\end{algorithm}

 In this section we describe the alternating least squares approach (summarized in Algorithm \ref{alg:opt_alt}) that is SVD free. This idea is based on the following
 key observation about the nuclear norm \cite{Hastie:14,srebro2004maximum} (see also a simple proof at \cite{sun:11}):
 \begin{align}
 \norm{X}_* = \min_{X = UV^\top} \frac{1}{2} (\norm{U}_F^2 +
 \norm{V}_F^2).
 \label{eq:mmmf}
 \end{align}
The paper \cite{Hastie:14} suggested combining the matrix factorization equality \eqref{eq:mmmf} with the
\textsf{SOFT-IMPUTE} approach in \cite{Mazumder:10} for matrix completion.

For our \textsf{TECROMAC} objective, instead of directly optimizing the nuclear norm, we can use the matrix factorization identity to solve the following ALM subproblem:
 \begin{align*}
 \{U^{r+1},V^{r+1},E^{r+1}\} = \arg\min_{U,V,E} L(U,V,E,Z^r,\mu^r),
 \end{align*}
 where
 \begin{align*}
 L(U,V,E,Z,\mu) = &\norm{P_{\Omega}(E)}_1 + \frac{\lambda_1}{2} \norm{U}_F^2 + \frac{\lambda_1}{2} \norm{V}_F^2
\\
& + \frac{\lambda_2}{2} \norm{UV^\top R}_F^2  + \langle Z, Y - UV^\top - E \rangle \\
& + \frac{\mu}{2} \norm{Y - UV^\top - E}_F^2.
\end{align*}
We alternate minimizing over $U,V$ and $E$ to optimize the objective
$L(U,V,E,Z,\mu)$. To update $U$ and $V$, we perform a gradient step:
\begin{align}
\label{eqn:alt_u}
U^{r+1} &= U^r - \eta \nabla_{U^r} L(U^r,V^r,E^r,Z^r,\mu^r)\\
\label{eqn:alt_v}
V^{r+1} &= V^r - \eta \nabla_{V^r} L(U^{r+1},V^r,E^r,Z^r,\mu^r),
\end{align}
where
\begin{align*}
\nabla_U L(U,V,E,Z,\mu) = &\lambda_1 U + \lambda_2 U V^\top R R^\top V \\
&- ZV - \mu(Y - UV^\top - E) V\\
\nabla_V L(U,V,E,Z,\mu) = &\lambda_1 V + \lambda_2 R R^\top V U^\top U \\
&- Z^\top U - \mu(Y - UV^\top - E)^\top U.
\end{align*}
After obtaining $U^{r+1}, V^{r+1}$, we update $E$ by substituting $X^{r+1}
= U^{r+1}(V^{r+1})^\top$ into \eqref{eqn:ipg_e}.

\begin{figure*}
\begin{center}
\animategraphics[width = 0.9 \textwidth,autoplay,loop,controls,poster=first]{1.0}{newfigures/escalator_}{3}{200}\\
\animategraphics[width = 0.9 \textwidth,autopause,loop,controls,poster=first]{1.0}{newfigures/lobby_}{3}{200}\\
\animategraphics[width = 0.9 \textwidth,autopause,loop,controls,poster=first]{1.0}{newfigures/watersurface_}{2}{200}
\caption{Comparisons of various approaches for cloud removal. 1, Original video. 2, Cloudy video. 3, Cloud detector. 4, Matrix Completion. 5, Robust Matrix Completion. 6, Interpolation. 7, \textsf{TECMAC}. 8, \textsf{TECROMAC}.}
\label{fig:simulation}
\end{center}
\end{figure*}

%-------------------------------------------------------------------------
\section{Simulation and quantitive assesment}
\label{sec:simulation}

\begin{figure*}[t]
\begin{center}
\animategraphics[width = 0.85 \textwidth,autoplay,loop,controls,poster=first]{1.0}{newfigures/modis_400_400_}{1}{180}
\caption{Cloud removal results on MODIS data. 1, Cloudy video. 2, Cloud detector. 3, MC. 4, Interpolation. 5, TECMAC. 6, \textsf{TECROMAC}.}
\end{center}
\label{fig:modis}
\end{figure*}

\begin{table}
\begin{center}
\begin{tabular}{c | c | c | c }
			Dataset & Escalator & Lobby & Watersurface \\
			\hline \hline
			MC & 0.2760 & 0.2677 & 0.9880 \\
			RMC & 0.2155 & 0.2465 & 0.9127 \\
			Interpolation & 0.0489 & 0.1151 & 0.0307 \\
			\textsf{TECMAC} & 0.0481 & 0.0646 & 0.0461 \\
			\textsf{TECROMAC} & \textbf{0.0153} & \textbf{0.0246} & \textbf{0.0045} \\
			\hline
\end{tabular}
\end{center}
\caption{Relative reconstruction error and times of various approaches}
\label{table:error}
\end{table}

\begin{table}
\begin{center}
\begin{small}
\begin{tabular}{c | c | c | c }
			Dataset & Escalator & Lobby & Watersurface \\
			\hline \hline
			Size & 62400 $\times$ 200 & 61440 $\times$ 200 & 61440 $\times$ 200 \\ \hline
			RRE(IPG) & 0.0153 & 0.0246 & 0.0045 \\
			RRE(ALT) & 0.0470 & 0.0219 & 0.0305 \\ \hline
			Time(IPG) & 147s & 441s & 156s \\
			Time(ALT) & 72s  & 145s & 52s  \\
			\hline
\end{tabular}
\end{small}
\end{center}
\caption{Comparison of accuracy and efficiency of two optimization algorithms to solve \textsf{TECROMAC}}
\label{table:time}
\end{table}

In this section we present extensive simulation results for our proposed algorithm. We manually added clouds to some widely used
background modeling videos\footnote{\url{http://perception.i2r.a-star.edu.sg/bk_model/bk_index.html}}. Since for the
simulation we have access to the ground truth $Y^*$ (i.e. the cloud-free
video) we can report the relative reconstruction error (RRE) for our estimation $\hat Y$ (shown in Table \ref{table:error}):
\begin{align*}
\textsf{RRE}(\hat Y, Y^*) = \frac{\norm{\hat Y - Y^*}_F^2}{\norm{Y^*}_F^2}.
\end{align*}

The thresholding parameter $\gamma$ was set to $0.6$, the regularization parameters $\lambda_1$ and $\lambda_2$ were set to be $20$ and $0.5$, respectively.  These parameter values were used for the experiments in the next section also.  The cloud removal and reconstruction results were shown in Figure \ref{fig:simulation} (best viewed in Adobe Reader). We make the following observations: 
\begin{itemize}
\item The cloud detection algorithm performs well and can handle the ``white background'' cases (e.g. some places in the escalator video) reasonably well; 
\item On some heavily cloud-contaminated frames, MC and RMC just output black frames. The temporally contiguous matrix completion approaches handles this issue
very well; 
\item \textsf{TECROMAC} slightly outperforms \textsf{TECMAC} (matrix completion with temporal continuity constraint added) in all cases, which verifies that using the robust $\ell_1$ loss is a preferred choice.
\end{itemize}

We also compare the performance and time efficiency of the propoposed computational algorithms, as summarized in Table \ref{table:time}, where for ALM-ALT algorithm we use rank $20$. As we can see, ALM-ALT generally is more efficient than ALM-IPG, while slightly sacrificing the recovery accuracy by adopting a non-convex formulation.

%

\if 0
\begin{figure*}
\begin{center}
\animategraphics[width = 0.85 \textwidth,autopause,loop,controls,poster=first]{1.0}{newfigures/brasil_1000_1000_}{1}{24}
\caption{Cloud removal results on LANDSAT-Brasil data.}
\end{center}
\label{fig:brasil}
\end{figure*}
\fi

%-------------------------------------------------------------------------
\section{Experiments on satellite image sequences}
\label{sec:real_data}

We also tested our algorithm on the real world MODIS satellite data\footnote{Publicly available \url{http://modis.gsfc.nasa.gov/data/}}, where we chose a subset consists of 181 sequential images of size $400 \times 400$. The results are shown in Figure \ref{fig:modis} (best viewed in Adobe Reader). We observe that
our algorithm recovers the background scenes from the cloudy satellite images very well, and visually produces much better recovery than existing models.

%------------------------------------------------------------------------
\section{Conclusion} \label{sec:conclusion} 
We have presented effective algorithms for cloud removal from satellite
image sequences. In particular, we proposed the ``\textsf{TECROMAC}'' (TEmporally Contiguous RObust MAtrix Completion)  approach to recover scenes of interest from partially
observed, and possibly corrupted observations.  We also suggested
efficient optimization algorithms for our model. The experiments demonstrated superior performance of the proposed methods on both
simulated and real world data, thus indicating our framework potentially
very useful for downstream applications of remote sensing with clean satellite
imagery.

\newpage 

{
\bibliographystyle{ieee}
\bibliography{egbib}
}

\end{document}